\documentclass[3p,twocolumn,times,sort&compress]{elsarticle}

\usepackage{multicol}
\usepackage{epstopdf} 
\usepackage{pdfpages}
\usepackage{array}
\usepackage{float}
\usepackage{amsmath,amssymb,amsfonts}
\usepackage{algorithmic}
\usepackage{textcomp}
\usepackage{graphicx,subfigure}
\usepackage{caption}
\usepackage{url}
\usepackage{soul}
\usepackage{xcolor}
\usepackage{tikz,xcolor,hyperref}
\allowdisplaybreaks

\definecolor{lime}{HTML}{A6CE39}
\DeclareRobustCommand{\orcidicon}{
	\begin{tikzpicture}
	\draw[lime, fill=lime] (0,0) 
	circle [radius=0.16] 
	node[white] {{\fontfamily{qag}\selectfont \tiny ID}};
	\draw[white, fill=white] (-0.0625,0.095) 
	circle [radius=0.007];
	\end{tikzpicture}
	\hspace{-2mm}
}
\foreach \x in {A, ..., Z}{%
	\expandafter\xdef\csname orcid\x\endcsname{\noexpand\href{https://orcid.org/\csname orcidauthor\x\endcsname}{\noexpand\orcidicon}}
}

\usepackage{amssymb}

\usepackage[figuresright]{rotating}

\begin{document}
\onecolumn {
\Large{
\begin{center}
 This work is published in IEEE Transactions on Medical Imaging.\\
\end{center}
P. Thiagarajan, P. Khairnar and S. Ghosh, "Explanation and Use of Uncertainty Obtained by Bayesian Neural Network Classifiers for Breast Histopathology Images," in IEEE Transactions on Medical Imaging, doi: 10.1109/TMI.2021.3123300.\\

\noindent Link to the paper can be found \href{https://ieeexplore.ieee.org/document/9585450}{\color{blue}here}.\\

\noindent © 2021 IEEE.  Personal use of this material is permitted.  Permission from IEEE must be obtained for all other uses, in any current or future media, including reprinting/republishing this material for advertising or promotional purposes, creating new collective works, for resale or redistribution to servers or lists, or reuse of any copyrighted component of this work in other works.
}}

\begin{frontmatter}



\title{Explanation and Use of Uncertainty Quantified by Bayesian Neural Network Classifiers for Breast Histopathology Images}


\author[1]{\fnref{lab1}Ponkrshnan Thiagarajan \orcidA{}}
\author[1]{\fnref{lab2}Pushkar Khairnar \orcidB{}}
\author[2]{\fnref{lab3} Susanta Ghosh \orcidC{}}
\fntext[lab1]{email: thiagara@mtu.edu}
\fntext[lab2]{email: ppkhairn@mtu.edu}
\fntext[lab3]{email: susantag@mtu.edu\\ This paper published in IEEE TMI.\\ DOI: 10.1109/TMI.2021.3123300}
\address[1]{Department of Mechanical Engineering-Engineering Mechanics, Michigan Technological University,MI,USA}
\address[2]{Mechanical  Engineering-Engineering Mechanics and the  Institute of Computing and Cybersystems,  Michigan Technological University, MI, USA. }
\begin{abstract}
Despite the promise of Convolutional neural network (CNN) based classification models for histopathological images, it is infeasible to quantify its uncertainties. Moreover, CNNs may suffer from overfitting when the data is biased. 
We show that Bayesian--CNN can overcome these limitations by regularizing automatically and by quantifying the uncertainty. We have developed a novel technique to utilize the uncertainties provided by the Bayesian--CNN that significantly improves the performance on a large fraction of the test data (about 6\% improvement in accuracy on 77\% of test data). 
Further, we provide a novel explanation for the uncertainty by projecting the data into a low dimensional space through a nonlinear dimensionality reduction technique. This dimensionality reduction enables interpretation of the test data through visualization and reveals the structure of the data in a low dimensional feature space.
 {We show that the Bayesian-CNN can perform much better than the state-of-the-art transfer learning CNN (TL-CNN) by reducing the false negative and false positive by 11\% and 7.7\% respectively for the present data set. It achieves this performance with only 1.86 million parameters as compared to 134.33 million for TL-CNN.} Besides, we modify the Bayesian--CNN by introducing a stochastic adaptive activation function. The modified Bayesian--CNN performs slightly better than Bayesian--CNN on all performance metrics and significantly reduces the number of false negatives and false positives (3\% reduction for both).  {We also show that these results are statistically significant by performing McNemar's statistical significance test. }
This work shows the advantages of Bayesian-CNN against the state-of-the-art, explains and utilizes the uncertainties for histopathological images. It should find applications in various medical image classifications.  
\end{abstract}

\end{frontmatter}
\section{Introduction}
\label{sec:introduction}
\subsection{Histopathological imaging for Breast cancer} 
 Breast cancer is the most common cause of cancer in women \cite{wild2014world,fitzmaurice2018global,bray2004changing}. It has the highest incidence (43.3 per 100 000 population) than any other cancer and the highest mortality rate (15\%) of all cancer deaths in women in 2012 \cite{wild2014world}. Thus detection and diagnosis of breast cancer are vital in reducing the impact of the disease. Histopathological imaging is considered the gold standard for breast cancer detection and diagnosis \cite{han2017breast}. Histopathology is a diagnostic technique that involves microscopic examinations of tissues to study the sign of a disease. This method preserves the underlying architecture of the tissues thereby providing a significant contribution to the diagnosis of diseases. It is the only way to detect some of the diseases like lymphocytic infiltration of cancer. Histopathological images have a lot of information and structure which makes them highly reliable in the diagnosis of diseases especially in almost all types of cancer \cite{gurcan2009histopathological}.
\subsection{state--of--the--art machine learning algorithms for Histopathological imaging} 
With the advancement in digital imaging, computer-aided diagnosis (CAD)  {is studied extensively in the literature } \cite{gurcan2009histopathological,ghaznavi2013digital,al2012digital,fuchs2011computational,giger2008anniversary,doi2007computer,aswathy2017detection}.  Machine learning algorithms such as conventional neural networks and deep neural networks have shown tremendous potential in CAD applications. Feature extraction in a conventional neural network played a significant role in the predictions of the neural network. Various methods of feature extraction followed by classifiers such as support vector machine, K-nearest neighbor, decision tree, Naive Bayes amongst others were explored in the literature for classification and segmentation of breast cancer images \cite{filipczuk2013computer,george2013remote,kowal2013computer,asri2016using,zhang2013breast,spanhol2015dataset,mercan2017multi}. Though these methods performed reasonably well in terms of accuracy, manual feature extraction was a bottleneck in improving the results.\\
\indent Convolutional neural networks (CNNs) were the most successful in learning advanced features automatically from the input images.  CNNs perform very well in classification, both binary \cite{bayramoglu2016deep,cruz2014automatic,spanhol2016breast} and multiclass classifications \cite{araujo2017classification,han2017breast}, and metastasis detection \cite{lin2019fast} on breast cancer images. Apart from being able to perform classification, CNNs were also successfully implemented in problems involving segmentation \cite{sirinukunwattana2016locality,badrinarayanan2017segnet,jia2017constrained} and detecting regions of interest that contain in-depth discriminatory information for classification in large whole-slide images \cite{li2019classification,yang2019guided,xu2019attention}.  CNNs also provided the scope to implement transfer learning. 
In transfer learning, the parameters learned for one data set are utilized to accelerate learning for another data set to solve a similar problem. 
Transfer learning CNN (TL-CNN) performs the best and is considered as the state-of-the-art classifier for breast cancer images  \cite{nawaz2018multi,motlagh2018breast,xie2019deep}. A grand challenge on breast cancer histology images was conducted to advance the state--of--the--art in classifying these images. It was concluded that CNN was the most successful method to classify these breast cancer images \cite{aresta2019bach}. The top performers \cite{chennamsetty2018classification,kwok2018multiclass,brancati2018multi} in this challenge used the architecture of an existing network such as Resnet \cite{he2016deep}, Densenet \cite{huang2017densely}, Inception \cite{szegedy2015going}, VGG16 \cite{simonyan2014very}, etc and pre-trained these networks using ImageNet \cite{krizhevsky2012imagenet}.\\
 \indent Uncertainty quantification provides a measure of trust in machine decisions and this metric is crucially important in medical applications  \cite{begoli2019need}. Given the success of automatic classifiers for medical images, their uncertainty quantification is essential. However, there are only a few studies that report uncertainty quantification \cite{gal2017deep}. Although CNN performs  {better} for image classification than other machine learning techniques in terms of accuracy, their parameters are deterministic and thus can not provide any measure of uncertainty in its predictions.  In addition, predictions based on deterministic CNNs might provide incorrect results and without any estimate of confidence, these might lead to undesirable consequences. To evaluate the confidence of predictions, uncertainty quantification is important \cite{begoli2019need}. The Bayesian CNN is an efficient state--of--the--art machine learning technique to quantify uncertainties \cite{blundell2015weight,shridhar2019comprehensive,shridhar2018uncertainty}. It endows the uncertainty quantification capability in the CNN framework with nominal additional computational effort. \\
 \indent In a Bayesian--CNN, the weights and biases are random variables as opposed to deterministic variables used in CNN. 
Through the stochastic nature of the parameters,  Bayesian--CNN captures the variability in the data set and computes the uncertainties in its predictions. 
Towards this, Blundell et al. \cite{blundell2015weight} introduced an efficient algorithm called ``Bayes by backprop". This algorithm can be used to learn the probability distribution on weights and it was compatible with the conventional backpropagation scheme. They have also shown that it is possible to regularise the weights by minimizing a loss function known as the variational free energy and demonstrated that this method of regularisation showed performance comparable with dropout on MNIST classification. Shridhar et al. \cite{shridhar2019comprehensive,shridhar2018uncertainty} extended the ``Bayes by backprop" algorithm to convolutional neural networks. The proposed Bayesian-CNN architecture was implemented for image classification, image super-resolution, and generative adversarial networks.  This method has shown performance as good as conventional CNN and in addition, provided uncertainty measures and regularisation. Kendall et al. \cite{kendall2017uncertainties} provided a Bayesian deep learning framework combining aleatoric and epistemic uncertainties. The authors improved the model's performance by 1 to 3\% over its deterministic counterpart by reducing the effect of noisy data. The trade--offs between modeling aleatoric and epistemic uncertainties were also studied. Kwon et al. \cite{kwon2018uncertainty} proposed a method of quantifying uncertainties in classification problems using Bayesian neural networks. Bayesian networks are used in the literature to perform active learning. Uncertainties were utilized to add new samples in the training data to improve performance by Gal et al \cite{gal2017deep}. An accurate, reliable active learning framework was introduced by Raczkowski et al \cite{rkaczkowski2019ara} utilizing the variational dropout-based uncertainty measures.  
\subsection{Present work:  Quantification, explanation, and use of uncertainty, and the Modified Bayesian--CNN} 
\indent Bayesian--CNNs are gaining popularity due to their ability to provide uncertainties associated with the predictions. Yet, there has been no known work to the authors' knowledge that demonstrates the advantages of Bayesian--CNN over CNN to classify histopathological images. In this work, at first, we perform uncertainty quantification for the classification of breast histopathological images and show the advantages of Bayesian--CNN over CNN. We found that the Bayesian--CNN improves accuracy and reduces overfitting in comparison to CNN in addition to quantifying uncertainties. To exploit the advantages of uncertainty quantification, we have developed a novel technique to divide the data into low and high uncertainty subsets. We show significant improvement in accuracy in the majority of data that has low uncertainty. A small fraction of the data that has higher uncertainty can be referred to the experts. Thus, uncertainty provides an avenue to determine which images require human intervention. We explain the reason for this high uncertainty in a subset of data by projecting the data into a low dimensional space (latent space) using t--distributed stochastic neighbor embedding (t-SNE). We choose a three-dimensional latent space to facilitate the visualization of data. To the best of the authors' knowledge, such an explanation of uncertainty is not reported in the literature.\\
\indent We propose a new model to extend the Bayesian--CNN to further improve its performance and name it as the modified  Bayesian--CNN. Adaptive activation functions have shown improvement in performance for both deep neural networks and convolutional neural networks \cite{qian2018adaptive,jagtap2020adaptive,dushkoff2016adaptive}. However such adaptive activation functions are not applied to Bayesian Neural networks.  The novelty of the proposed adaptivity lies in the fact that the learnable parameter is random as opposed to deterministic parameters reported in the literature. Thus the activation function is now sampled from an ensemble of such functions based on the probability distribution of parameters learned from the data.\\
\indent The rest of the paper is organized as follows: in section \ref{CHAPTER2} the data set and the methods used in this work are described; results obtained in the work and their analysis are presented in section \ref{CHAPTER3} followed by conclusions in section \ref{CHAPTER4}.   

\section{Data set and methods}\label{CHAPTER2}
\subsection{Data acquisition}
The breast histopathological images used in this work are from a publicly available data set \cite{mooney_b_2017}. These are images containing regions of Invasive Ductal Carcinoma  which is the most common subtype of all breast cancers \cite{mooney_b_2017}. These regions are separated from the whole slide images by pathologists. The original data set consisted of 162 whole mount slide images of Breast Cancer specimens scanned at 40x. 
From the original whole slide images, 277,524 patches of size $50\times50$ pixels were extracted (198,738  negative and 78,786 positive) and provided as a data set for classification.

\subsection{Data Preparation} \label{data_prep}
The data set consists of two classes: positive (1) and negative (0). The classification of the data was carefully done by experienced pathologists providing the ground truth for training. 20\% of the entire data set was used as the testing set for our study. The remaining 80\% of the entire data was further split into a training set and a validation set (80\%-20\% split) to perform hyperparameter optimization.  

The image size is $3\times50\times50$ (D$\times$H$\times$W), where D is the depth (color channels), H is the height, and W is the width. The images were shuffled and converted from uint8 to float format for normalizing.
For most of the images, the majority of the pixel values are greater than 200. 
This makes the computation extremely expensive especially with the large size of data. Further, it leads to problems such as singularity/gradient explosion during the evaluation of loss function. In addition, the original images are such that the information that needs to be highlighted has low pixel values. To solve these problems, we computed the complement of all the images (training and testing) and then used them as inputs to the neural network.  {The pixel-wise normalization and complement was carried out as $p_n = (255-p)/255 $. Where $p$ is the original pixel value and $p_n$ is the pixel value after normalization and complement.}
\subsection{Methods}\label{sec:method}
 The mathematical background for Bayesian--CNN and its uncertainty quantification \cite{shridhar2019comprehensive}, and for t-SNE \cite{van2008visualizing} is revisited in this section. The implementation of stochastic adaptive activation on Bayesian--CNN is also described. 

\subsubsection{ {Bayesian Neural Networks}} \label{sec:BNN}
The neural network can be defined as a probabilistic model ${P}({y}|{x}, {w})$, where ${x} \in \mathbb{R}^{{p}}$ is an input to the network, ${y} \in {\Upsilon}$ is the output. Where $\Upsilon$ is a set of all possible outputs. The network consists of a set of trainable random parameters $w$. This set of parameters $w$ is learned using a complete Bayesian approach. In this approach given training data $D$, the posterior distribution of weights ${P}({w} | {D})$ is calculated using Bayesian inference for neural networks which involves marginalization over all possible values of $w$. Once the posterior distribution ${P}({w} | {D})$ is obtained, predictions on the unseen data are obtained by taking expectations on the predictive distributions. The predictive distribution of an {unknown label} ${\widehat{y}}$ of a {test data item} ${\widehat{x}}$ is given by
$
{P}(\widehat{{y}} | \widehat{{x}})=\mathbb{E}_{{P}({w} | {D})}[{P}(\widehat{{y}} | \widehat{{x}}, {w})]
=\int_{\Omega_{w}} {P}(\widehat{{y}} | \widehat{{x}}, {w}) {P}({w} | {D}) {dw}.\\
$
\indent To estimate the posterior distribution we use Bayes' Rule which gives:
\begin{align}
{P}({w} | {D})=\frac{{P}({D} | {w}) {P}({w})}{{P}({D})}
\label{eq:bayes}
\end{align}
The term ${P}({D} | {w})$ is the likelihood of the training data (${D}$) given a parameter setting (${w}$). Assuming each training data to be independent and identically distributed, the above term becomes the product of likelihood, 
$
{P}({D} | {w})=\prod_{{n=1}}^{{N}} {P}\left({y}^{n} | {w}, {x}^{n}\right)
$ where, $({x^n, y^n})$ are the training data item and known label respectively.\\
 The prior ${P}({w})$ is our belief about the distribution of weights without seeing the data. The term ${P}({D})$ in the equation (\ref{eq:bayes}) is intractable which makes the posterior distribution ${P}({w} | {D})$ intractable. The term ${P}({D})$ involves marginalization over the weight distribution: $
{P}({D})=\int_{{\Omega_{w}}} {P}({D} | {w}) {P}({w}) {dw}
$.\\
\indent For variational inference \cite{blundell2015weight}, the posterior distribution ${P}({w} | {D})$ which is intractable is approximated with a tractable simpler distribution over the model weights ${q}({w})$. If we assume our weight distributions to be Gaussian, this simpler tractable distribution has variational parameters ${\theta}$, where ${\theta}=({\mu}, {\sigma})$. We fit the variational parameters ${\theta}$ such that ${q}({w} | {\theta}) \approx {P}({w} | {D})$. The variational posterior ${q}({w} | {\theta})$ is used instead of the intractable posterior ${P}({w} | {D})$ for the inference,
$
{P}(\widehat{{y}} | \widehat{{x}})=\mathbb{E}_{{q}({w} | {\theta})}[{P}(\widehat{{y}} | \widehat{{x}}, {w})]
$. Estimating ${q}({w} | {\theta}) \approx {P}({w} | {D})$ we can say that we have learned the distribution of weights given training data.

To make the variational posterior ${q}({w} | {\theta})$ and the true posterior ${P}({w} | {D})$ similar, we minimize the KL divergence between them:
\begin{align}
&{K} {L}[{q}({w} | {\theta}) \| {P}({w} | {D})]=\mathbb{E}_{{q}({w} | {\theta})}\left[{\log} \frac{{q}({w} | {\theta})}{{P}({w} | {D})}\right] \notag \\
&=\int {q}({w} | {\theta}) {\log} \frac{{q}({w} | {\theta})}{{P}({w})} {d} {w} +\int {q}({w} | {\theta}) {\log} ({P}({D})) {d} {w} \notag\\&-\int {q}({w} | {\theta}) {\log} ({P}({D} | {w}))  {dw}
\end{align}
The term ${\log} ({P}({D}))$ makes the above equation intractable. Although intractable, it is constant. Therefore, minimizing the KL divergence can be defined as:
\begin{align}
{\min}\left ({K} {L}[{q}({w} | {\theta}) \| {P}({w} | {D})]\right) &={\min} ({K} {L}[{q}({w} | {\theta}) \| {P}({w})]\notag\\&-\mathbb{E}_{{q}({w} | {\theta})}[\log ({P}({D} | {w}))]
\end{align}
The term $\left({K} {L}[{q}({w} | {\theta}) \| {P}({w})]
-\mathbb{E}_{{q}({w} | {\theta})}[\log ({P}({D} | {w}))]\right)$ is called the {Variational Free Energy (VFE)} which is to be minimized. 
Exact minimisation of the cost function is computationally expensive, therefore it is approximated through a Monte Carlo sampling procedure as follows:
\begin{align}
{\mathcal{F}}({D}, {\theta}) &\approx \sum_{i=1}^{n} [{\log} {q}\left({w}^{(i)} | {\theta}\right)-{\log} {P}\left({w}^{(i)}\right)-{\log} {P}\left({D} | {w}^{(i)}\right)]
\label{eq:loss}
\end{align}
Where, ${w}^{(i)}$ denotes the $i^{th}$ Monte Carlo sample drawn from the variational posterior ${q}({w}^{(i)} | {\theta})$.  {Each term in ${\mathcal{F}}({D}, {\theta})$, the loss function that needs to be minimized, also known as the cost function has weights that are drawn from the variational posterior. For further details on the cost function, readers are referred to} \cite{blundell2015weight,shridhar2019comprehensive}. 
To implement and compute the above cost we need three terms: 1) log of variational posterior, 2) log of prior (Gaussian or scale mixture), 3) log-likelihood of the data.\\
\indent The variational posterior ($q$) is composed of independent Gaussian distribution for each parameter. The sample of weights are obtained by sampling the unit Gaussian, shifting and scaling by the mean ${\mu}$ and a standard deviation ${\sigma}$ respectively. To ensure that the standard deviation is always non-negative, it is expressed as $\sigma=\operatorname{softplus}{(\rho)=\log (1+\exp (\rho))}$, point-wise. Parameters, $w$, of the variational posterior, $q$, can be re-written in terms of a parameter--free noise as ${[w=\mu+\log (1+\exp (\rho)) \circ \varepsilon]}$ or ${[w=\mu+ \sigma \circ \varepsilon]}$.  Where, $\circ$ is point-wise multiplication, $\varepsilon$ is a sample drawn from unit normal \cite{blundell2015weight}. The steps for optimization are presented in supplementary document.\\


\subsubsection{Bayesian Convolution Neural Networks}
\begin{figure}[H]
    \centering
        \includegraphics[scale=0.25]{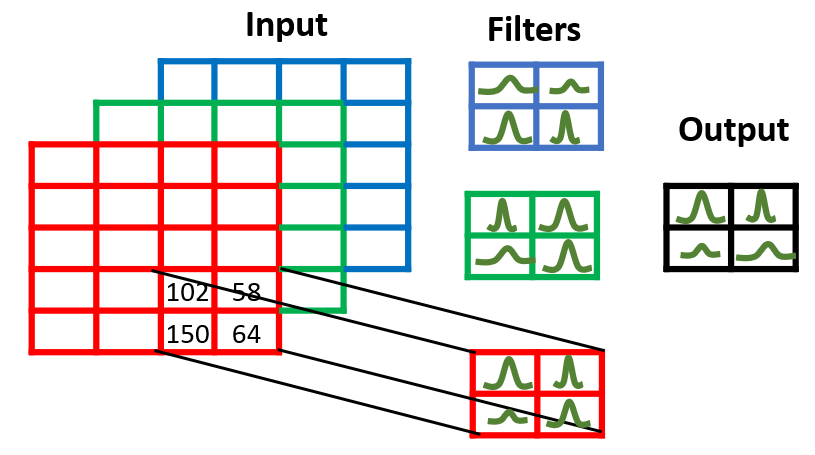}
    \caption{Schematics of the convolution of Bayesian--CNN showing probability distributions of random parameters.}
    \label{fig:conv}
    \vspace{-10pt}
\end{figure}

The parameters of CNNs are filters or kernels which are to be learned during training. In the case of a Bayesian--CNN these kernels are represented by probability distributions as shown in Fig.~\ref{fig:conv}.

During the training of Bayesian--CNN the reparameterization trick is applied on these filters or kernels, which are of shape {h$\times$w$\times$d}. They are sampled from the variational posterior ${q}({w} | {\theta})$ using the following equation:
\begin{equation}
  w_{h, w, d}=\mu_{h, w, d}+\log \left(1+\exp \left(\rho_{h, w, d}\right)\right) \circ \varepsilon_{h, w, d}  
\end{equation}

where h is the height, w is the width and d is the depth of the filter and $\circ$ represents point-wise multiplication. After the sampling from the variational posterior, similar optimization steps as explained in the previous section are followed.
\subsubsection{Uncertainty Quantification}
\label{sec:UQ}
The uncertainty quantification becomes extremely important when dealing with the applications related to autonomous vehicles, medical imaging, etc. Bayesian deep learning makes it possible to quantify the uncertainties in the prediction as we have probability distribution over weights. Taking an expectation of the predictive posterior probability distribution: $\mathbb{E}_{{q}({w} | {\theta})}[{P}(\widehat{{y}} | \widehat{{x}}, {w})]$ gives us the most probable prediction of the unknown data $\widehat{{x}}$. The variance of the predictive posterior probability distribution: $\operatorname{Var}_{{q}({w} | {\theta})}[{P}(\hat{{y}} | \hat{{x}}, {w})]$ quantifies the uncertainties. There are two types of uncertainties: the {Aleatoric} and the {Epistemic} uncertainty. The variance of the predictive posterior probability distribution is the sum of both these uncertainties:
\begin{align}
\operatorname{Var}_{q(w | \theta)}[P(\widehat{y} | \widehat{x}, w)]
&= \textrm{aleatoric} + \textrm{epistemic}
\end{align}
The aleatoric uncertainty corresponds to the noise in the data set whereas the epistemic uncertainty corresponds to the variability of the model developed\cite{shridhar2019comprehensive}. One of the promising approaches to quantify the uncertainties is explained in \cite{kwon2018uncertainty}. The uncertainties are obtained from the variance of the predictive posterior probability distribution,
\begin{align}
&\operatorname{Var}_{{q}({w} | {\theta})}[{P}(\widehat{{y}} | \widehat{{x}}, {w})]=\mathbb{E}_{{q}({w} | {\theta})}\left[({y}-\mathbb{E}[{y}])^{2}\right] \notag \\
 &=\int_{\Omega_{w}} \left[\left[\operatorname{diag}\left(\mathbb{E}_{P(\widehat{{y}} | \widehat{{x}}, {w})}[\widehat{{y}}]\right) \right. \right.  \left.-\mathbb{E}_{P(\widehat{y} | \widehat{x}, w)}[\widehat{y}] \mathbb{E}_{P(\widehat{y} | \widehat{x}, w)}[\widehat{y}]^{T}\right] \notag\\ & \left.q(w | \theta) dw \right] +\int_{\Omega_{w}} \left[\left[\mathbb{E}_{P(\widehat{y} | \widehat{x}, w)}[\widehat{y}]-\mathbb{E}_{q_{\theta}(\widehat{y} | \widehat{x})}[\widehat{y}]\right]\right.
 \notag\\ & \left.\left[\mathbb{E}_{P(\widehat{y} | \widehat{x}, {w})}[\widehat{y}] - \mathbb{E}_{{q}_{{\theta}}(\mathcal{Y} | {x})}[\widehat{{y}}]\right]^{T} {q}({w} | {\theta}) d {w} \right] \label{eq:var}
\end{align}
The above expression is the sum of aleatoric and the epistemic uncertainties which is derived from the variant of the law of the total variance \cite{kwon2018uncertainty}

The first term of Eq.~\ref{eq:var} is defined as the aleatoric uncertainty and the second term is the epistemic uncertainty. Due to the integral term in Eq.~\ref{eq:var}, it is intractable and requires approximations as follows. The equation of the aleatoric uncertainty is defined as:
$
{\frac{1}{N} \sum_{n=1}^{N}} \operatorname{diag}\left(\widehat{{p}}_{n}\right)-\widehat{{p}}_{n} \widehat{{p}}_{n}^{T} 
\text{ where, }\widehat{{p}}_{{n}}={p}\left(\widehat{{w}}_{{n}}\right)=\operatorname{softmax}\left\{{f}^{\widehat{w}_{t}}(\widehat{{x}})\right\}
$. And the equation of the epistemic uncertainty is defined as:
$
{\frac{1}{N} \sum_{n=1}^{N}}\left(\widehat{{p}}_{{n}}-{\overline { p }}\right)\left(\widehat{{p}}_{{n}}-\overline{{p}}\right)^{T}
\text{where, } \widehat{{p}}_{n}={p}\left(\widehat{{w}}_{{n}}\right)=\operatorname{softmax}\left\{{f}^{\hat{W}_{t}}(\widehat{{x}})\right\} \text{and } \overline{{p}}=\sum_{n=1}^{N} \frac{\widehat{{p}}_{{n}}}{{N}}
$.
The overall equation for the variance is:
\begin{align}
\operatorname{Var}_{q(w | \theta)}[P(\widehat{y} | \widehat{x}, w)]&=\left(\frac{1}{N} \sum_{n=1}^{N} \operatorname{diag}\left(\widehat{p}_{n}\right)-\widehat{p}_{n} \widehat{p}_{n}^{T}\right)
\notag\\&+\left(\frac{1}{N} \sum_{n=1}^{N}\left(\widehat{p}_{n}-\bar{p}\right)\left(\widehat{p}_{n}-\bar{p}\right)^{T}\right)
\label{eq:var_approx}
\end{align}

The variance of the predictive distribution can be calculated by Eq.~\ref{eq:var_approx} which provide us with the confidence of the network in making predictions for a given image.

\subsubsection{Adaptive activation }
The weights and bias of a perceptron in a neural network perform a linear transformation of the inputs. The output of this linear transformation is passed to an activation function that determines if a particular perceptron is activated for a given input. A non-linear activation function is a key component of a neural network, which enables it to learn complex functions with a small number of perceptrons. However, this nonlinearity of the activation function is known to introduce problems such as exploding or vanishing gradients. Thus, there is a trade-off between the learning capabilities vs training complexities of a perceptron, which can be optimized. To achieve this, we modify the Bayesian-CNN by introducing a learnable stochastic activation function that adapts to the training data.
Such adaptivity in the context of other machine learning models has been reported in the literature \cite{qian2018adaptive,dushkoff2016adaptive}, in which the learnable parameter is deterministic. In this work, we introduce a probabilistic learning parameter for the adaptivity of Bayesian-CNN.\\
This adaptive activation function contains a probabilistic parameter that is learned during the training of the neural network. To this end, a trainable probabilistic hyperparameter ($\alpha$) is introduced in the activation functions of a Bayesian--CNN and the resulting network is the modified Bayesian--CNN. The details of the proposed adaptive activation function are given below.
\begin{equation}
    \sigma(\alpha f_k(x^{k-1}))
\end{equation}
    
 where\\
$
f_k(x^{k-1}) = \overline{w}^k x^{k-1} + b^k
$\\ $\sigma$ is the activation function\\ $\alpha$ is the probabilistic hyperparameter that can be trained \\  $\overline{w}$ and $b$ are the weights and bias of the $k^{th}$ layer and $x^{k-1}$ is the output from the previous layer of the neural network.\\
\par {To incorporate the stochastic adaptive activation, the loss function $\mathcal{F}$ (Eq.}~\ref{eq:loss}{) is modified to include the additional stochastic parameter $\alpha$. Specifically, the set of trainable network parameters $w = \{\overline{w}, b\}$ in $\mathcal{F}$  is extended to $W = \{\overline{w}, b,\alpha\}$. Thus the modified loss function $\widetilde{\mathcal{F}}(D,\theta)$ is given as}
\begin{align*}
\widetilde{\mathcal{F}}({D}, {\theta}) &\approx \sum_{i=1}^{n} [{\log} {q}({W}^{(i)} | {\theta})-{\log} {P}({W}^{(i)})-{\log} {P}({D} | {W}^{(i)})]
\label{eq:loss_mod}
\end{align*}
{The parameter $\alpha$ is learned in the same way as the parameters $w$ as described in Sec.}~\ref{sec:BNN}{. A prior distribution (a mixture of Gaussian) for the parameter $\alpha$ is assumed and the posterior distribution is obtained via Bayes' rule. This posterior distribution is approximated by a Gaussian variational posterior $q_\alpha$  with mean $\mu_\alpha$ and standard deviation $\sigma_\alpha$. The parameters of this variational posterior are updated through gradient descent as,\\
$\mu_\alpha^{m+1} = \mu_\alpha^m - \eta\, \nabla_{\mu_\alpha} \widetilde{\mathcal{F}}^{\,m}$ and\\ $\sigma_\alpha^{m+1} = \sigma_\alpha^m - \eta \,\nabla_{\sigma_\alpha} \widetilde{\mathcal{F}}^{\,m}$
\\Where, $\eta$ is the learning rate. Refer to section II of the supplementary document for further details.  }\\

In this work, an adaptive ReLu given by  {$\sigma (\gamma) = \max (0,\gamma x) $} is used for the fully connected layers.\\

The networks were trained, validated, and tested on GPUs using Apache MXNet on Python 3. Details of the computational platform are given in the supplementary document.
\subsubsection{t-distributed stochastic neighbor embedding (t-SNE)}
t-SNE \cite{van2008visualizing} is an unsupervised non-linear dimensionality reduction technique that helps in visualizing high dimensional data by mapping it into a low dimensional space. In this method, as a first step, a conditional probability ($p_{j/i}$) is assigned between data points $x_i$ and $x_j$ in a high dimensional space.
This conditional probability $p_{j/i}$ represents the similarity of data points such that similar points in the high dimensional space have a higher probability than dissimilar points. One such measure of similarity is the  {Euclidean} distance between a pair of data points.  In the second step, a similar conditional probability ($q_{j/i}$) is defined for the low dimensional counterpart $y_i$ and $y_j$. If the low dimensional representation of the data correctly models similarity between the points in the high dimensional space, the two conditional probabilities $p_{j/i}$ and $q_{j/i}$ will be equal. Thus, in the third step, by minimizing the KL divergence between the two conditional probabilities a low dimensional map of the data points is obtained via t-SNE.
In this work, we explain the high uncertainty of images by projecting  {them} into a latent space via t-SNE. 

\section{Results and analysis}\label{CHAPTER3}
This section presents classification results for histopathological data, its uncertainty quantification obtained through the proposed modified Bayesian--CNN (Section \ref{sec:method}). It also provides an explanation of the predicted uncertainty via t-SNE.
Further, the proposed model is analyzed and juxtaposed with Bayesian--CNN and transfer learning CNN (TL-CNN). 

\subsection{Hyperparameter optimization}
\begin{figure}[H]
    \centering
   \includegraphics[width=0.5\textwidth]{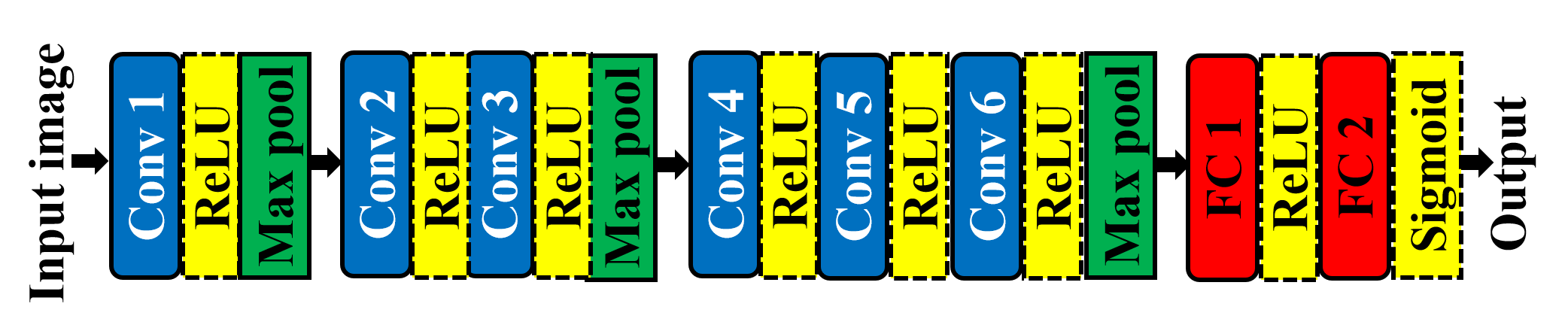}
    \caption{Architecture for Bayesian--CNN and modified Bayesian--CNN. Conv represents convolution layers and FC represent fully connected layers. Number of neurons in FC 2 is fixed as 2 for binary classification.  {Optimized} kernel size of all convolution layers is fixed as $3\times 3$ with stride 1.  {Optimized} kernel size of max pool layers is $2 \times2$ with stride 2. }
    \label{fig:arch}
\end{figure}

Hyperparameters and architecture for all the networks considered here are chosen through hyperparameter optimization. 
A Tree-structured Parzen Estimator (TPE) algorithm is used which is a sequential model-based optimization approach \cite{bergstra2011algorithms}. In this approach, models are constructed to approximate the performance of hyperparameters based on historical measurements. New hyperparameters are chosen based on this model to evaluate performance. A python library Hyperopt \cite{bergstra2013making} is used to implement this optimization algorithm over a given search space. Data is split into a training set and a validation set as explained in the Data preparation section \ref{data_prep}. An optimization is performed to maximize the validation accuracy for different configuration and hyperparameter settings of the network. \\
\indent The architecture of TL-CNN follows VGG-16 \cite{simonyan2014very}. Architectures for Bayesian--CNN and modified Bayesian--CNN are shown in Fig.~\ref{fig:arch}. {In the case of modified Bayesian--CNN, the ReLU function after the FC 1 layer (see Fig.}\ref{fig:arch}{) is changed into a stochastic--adaptive ReLU. Using stochastic-adaptive Relu in convolution layers as well increased the computational cost without any significant improvement in performance. More details are given in the supplementary document.} The number of filters for convolution layers and the number of neurons in the fully connected layers are obtained through hyper-parameter optimization. The results of the optimisation are presented in Tables.~\ref{Tab:arch} and \ref{Tab:arch1}. 

\begin{table}[H]
    \centering
    \caption{Selected hyperparameters}
    \begin{tabular}{ |p{3cm}|p{1.2cm}|p{1.2cm}|p{1.2cm}|} 
        \hline
        \textbf{Network} & \textbf{Learning rate} & \textbf{Batch Size} & \textbf{Layers learnt}\\
        \hline
        TL-CNN & 0.001 &64 & 2\\  \hline
        Bayesian--CNN &0.001 & 128  & All\\ \hline
        Modified Bayesian-CNN &0.0001 &64 & All \\ 
        \hline
    \end{tabular}
    \label{Tab:arch}
    \vspace{-10pt}
\end{table} 

\begin{table}[H]
    \centering
    \caption{Selected hyper parameters for network architecture} \vspace{-3pt}
    \begin{tabular}{ |p{1.4cm}|p{0.5cm}|p{0.5cm}|p{0.5cm}|p{0.55cm}|p{0.5cm}|p{0.5cm}|p{0.4cm}|p{0.4cm}|} 
        \hline
        \textbf{Network} & \textbf{con1} & \textbf{con2} & \textbf{con3} & \textbf{con4}& \textbf{con5}& \textbf{con6}& \textbf{FC1}\\
        \hline
        Bayesian--CNN &16 & 32  & 32 & 64 & 128 & 256 & 512 \\ \hline
        modified Bayesian-CNN &32 &64 &64 &128 &128 &128 &256\\ 
        \hline
    \end{tabular}
    \label{Tab:arch1}
    \vspace{-10pt}
\end{table}

\subsection{Training, validation, and testing of networks}
The networks are trained using the images as inputs and the corresponding known classes as outputs. The cross-entropy loss and the variational free energy loss are minimized for TL-CNN and Bayesian networks respectively. 
The training is done until the loss converges or the validation accuracy starts to decrease. In the following the details of training, validation, and testing for TL-CNN, Bayesian--CNN, and the modified Bayesian--CNN are described.  
\subsubsection{Convolutional neural networks}
\begin{figure}[H]
    \centering
   \includegraphics[width=0.5\textwidth]{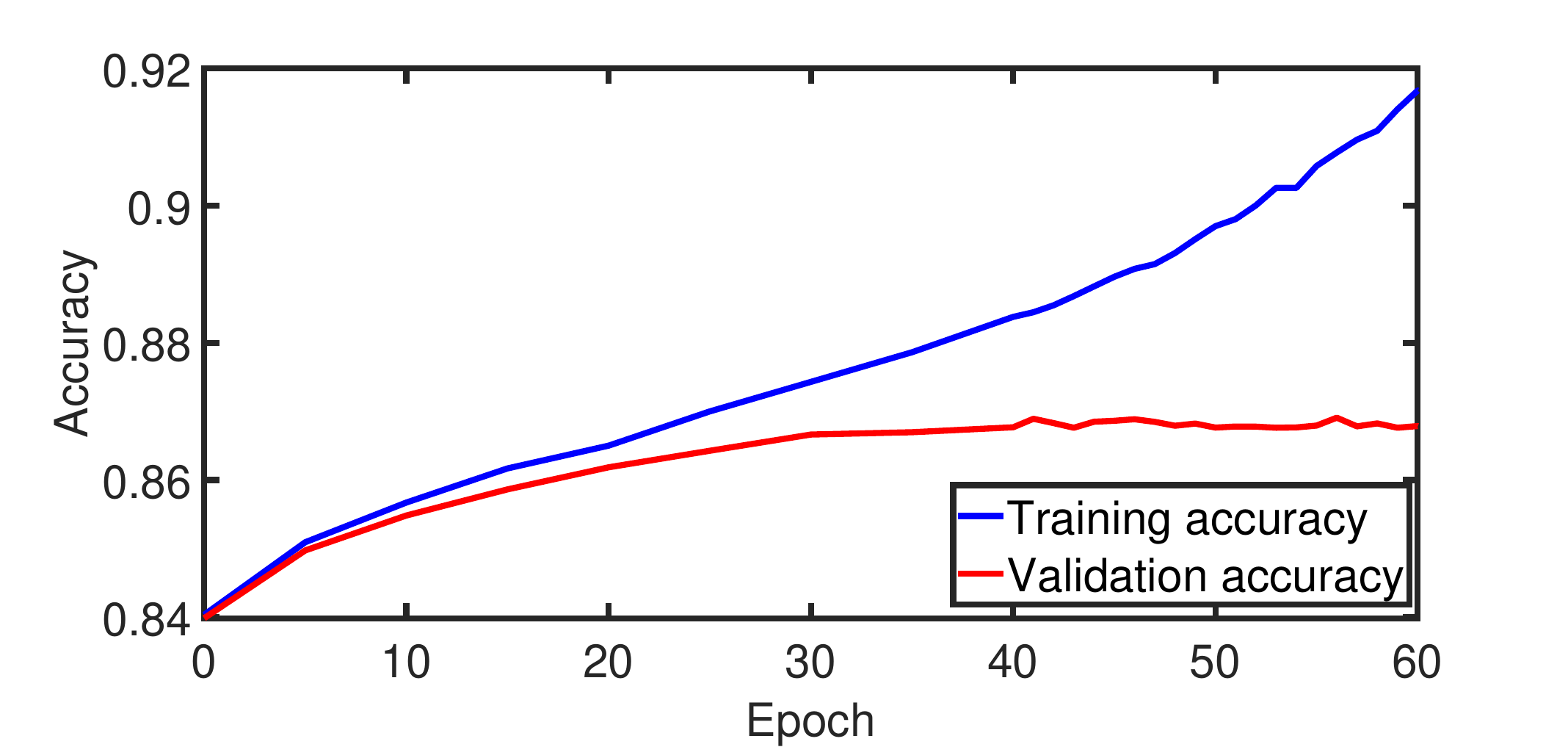}
    \caption{Training and validation accuracy of TL-CNN. We can clearly see the problem of overfitting as the training accuracy increases while the validation accuracy does not improve. The results presented for the TL-CNN henceforth are for the maximum validation accuracy.}
    \label{fig:acc_cnn}
\end{figure}

Transfer learning is the state-of-the-art for CNNs and has consistently performed the best for medical image classification \cite{aresta2019bach,shin2016deep}. 
In transfer learning, the learned parameters for one data set {are} utilized for other data sets to perform a similar task.  
Here we use the weights of VGG16 \cite{simonyan2014very} architecture trained on Imagenet data set \cite{krizhevsky2012imagenet} for classification and perform transfer learning to classify the histopathological data.
The accuracy for training and validation for each epoch is shown in Fig.~\ref{fig:acc_cnn} for TL-CNN. The accuracy increases with the epoch for the training data whereas the accuracy remains constant for the validation data after the first few epochs.
This problem is known as overfitting. Thus, TL-CNN fails to generalize its predictions to unknown validation data and overfits the training data.
Beyond 40 epochs, the validation accuracy remains almost a constant while the training accuracy is still increasing, which is long before the training loss has converged. 

To alleviate the overfitting of the networks, regularization strategies are used, such as $L_1$ or $L_2$ regularization, cross-validation, early stoppage, and dropout \cite{bishop2006pattern, orr2003neural, friedman2001elements}. However, these may require human intervention during training (such as parameter tuning, determination of early stoppage) or may lead to loss of information due to drop out \cite{orr2003neural}.  \\
{In Bayesian networks the prior distribution on weights $p(w)$ introduces regularisation of weights automatically. A Gaussian prior yields $L_2$ regularisation on the weights and a Laplace prior yields $L_1$ regularisation on the weights. In addition, due to the stochastic nature of the parameters, an average across multiple models is computed during training which introduces a regularisation effect on the network. However, similar to the selection of a regularisation parameter in deterministic networks, the selection of prior parameters in Bayesian networks is challenging.}
Herein we show that through a Bayesian framework the limitations of overfitting is overcome automatically.\\ 

\subsubsection{Bayesian--CNN and the modified Bayesian--CNN} 
\begin{figure}[H]
    \centering
    \includegraphics[width=0.5\textwidth]{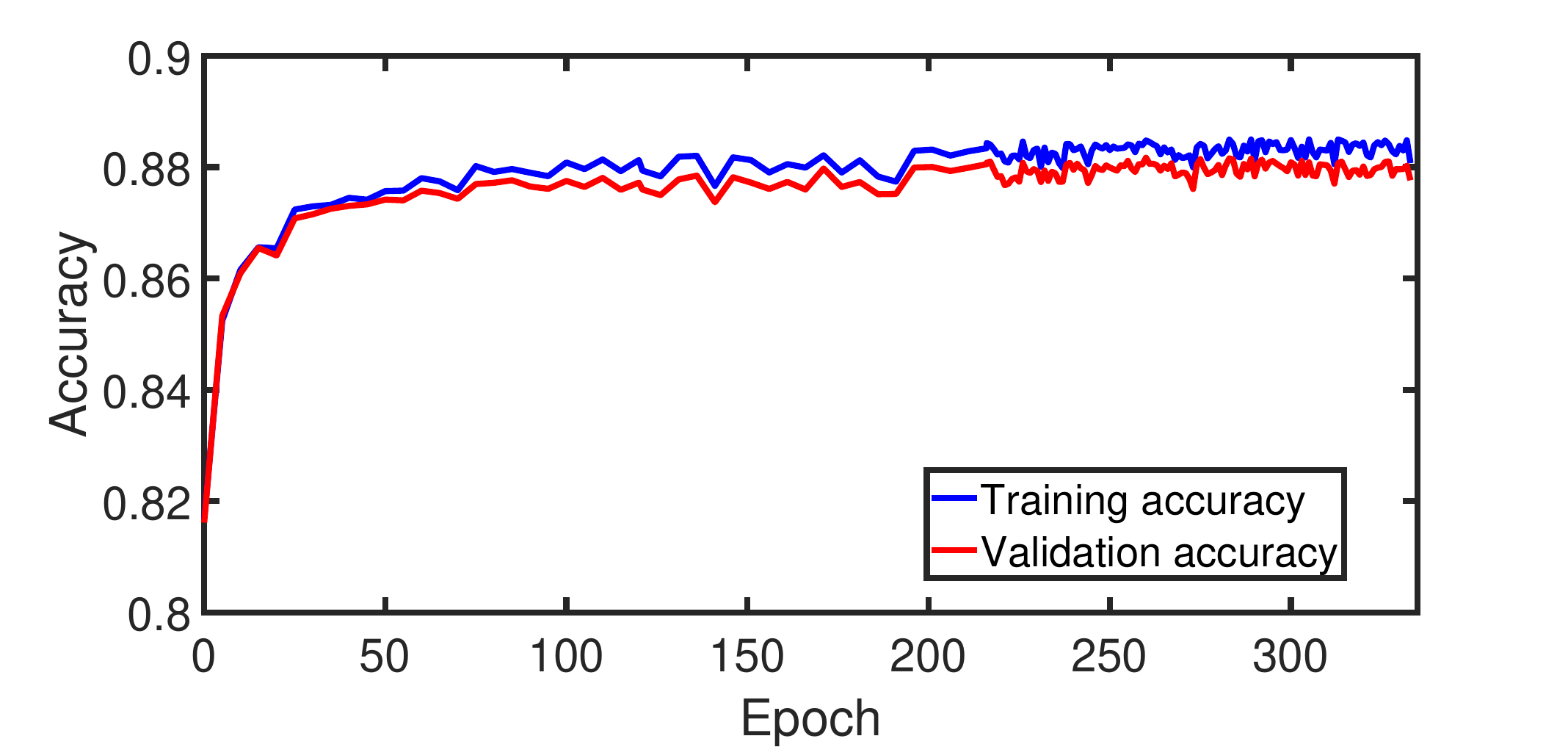}
    \caption{Training and validation accuracy of Bayesian--CNN.  We can see that both training and validation accuracy increases with epoch which implies automatic regularisation.}
    \label{fig:acc_bcnn}
\end{figure}

\begin{figure}[H]
    \centering
    \includegraphics[width=0.5\textwidth]{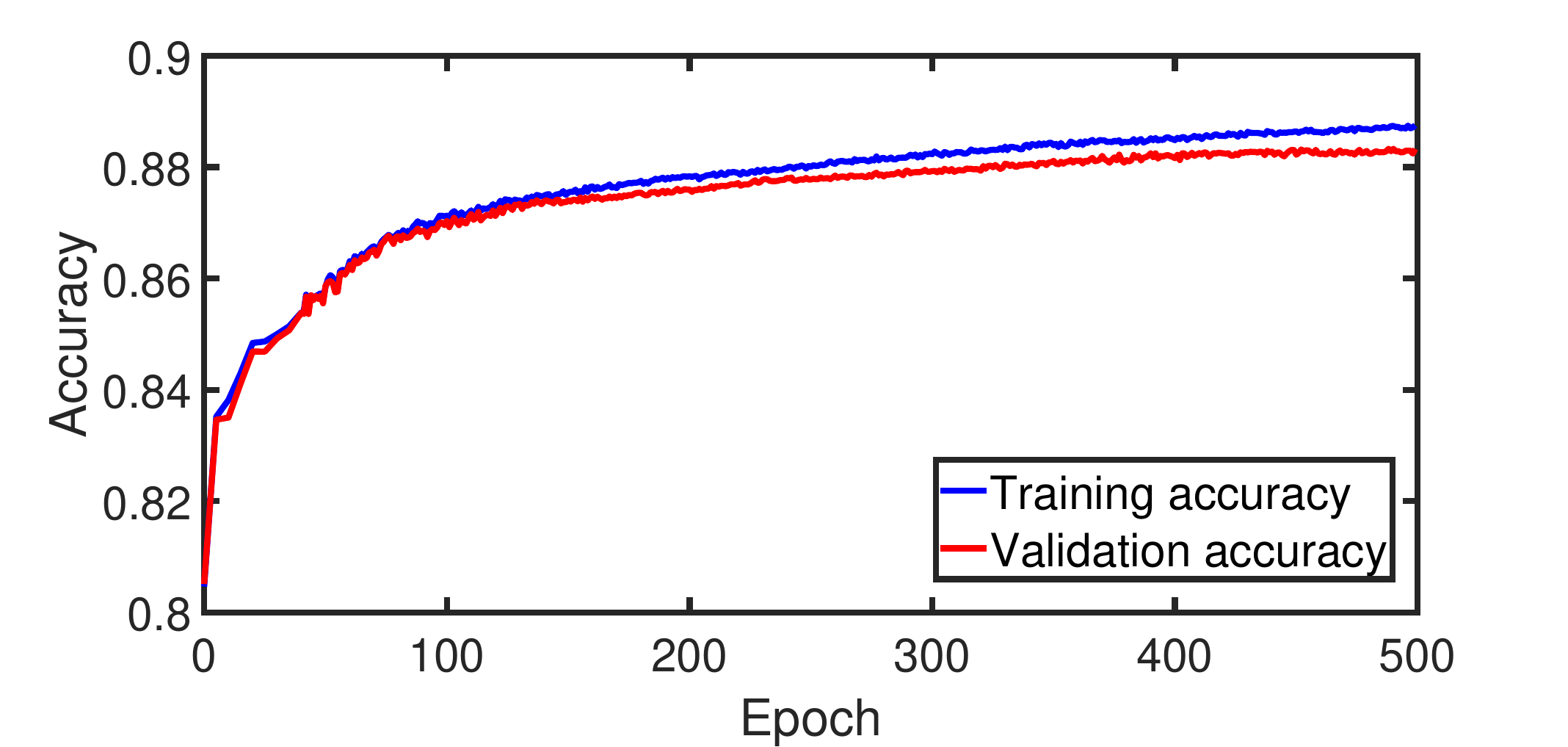}
    \caption{Training and validation accuracy of modified Bayesian--CNN.}
    \label{fig:acc_mbcnn}
\end{figure}

The accuracy of predictions for both training and validation data of the Bayesian and modified Bayesian-CNN is shown in Fig.~\ref{fig:acc_bcnn} and Fig.~\ref{fig:acc_mbcnn} respectively. It is evident that the problem of overfitting is eliminated by using a Bayesian network as the accuracy of predictions for both training and validation data increases with an increase in the number of epochs. 

The values of the accuracy of training, validation, and test data upon the convergence of the loss function are provided in Table \ref{Tab:comp1}. It is seen that the problem of overfitting drastically reduces the accuracy for validation and test data in a TL-CNN whereas this problem is not present for Bayesian--CNN and the modified Bayesian--CNN. The training, validation, and test accuracies of the modified Bayesian--CNN is slightly higher than the Bayesian--CNN when the loss function is converged. 
We found that the modified Bayesian--CNN requires about the same time as that of Bayesian--CNN for the same batch size but twice the computational cost per epoch during the training as compared to TL-CNN. More details on the training times are presented in the supplementary document.   {We have performed a comparative study of the Bayesian--CNN and the modified Bayesian--CNN on the MNIST data \cite{lecun1998mnist}. We found that the modified Bayesian--CNN shows a similar improvement in performance on the accuracy for the MNIST data set. Further details are presented in the supplementary document. }
\begin{table}[H]
    \centering
    \caption{A Comparison of TL-CNN, Bayesian--CNN and modified Bayesian-CNN}
    \begin{tabular}{ |p{3cm}|p{1.2cm}|p{1.2cm}|p{1.2cm}|} 
        \hline
        \textbf{Network} & \textbf{Training Accuracy} & \textbf{Validation Accuracy} & \textbf{Testing Accuracy}\\
        \hline
        TL-CNN & 0.9078 &0.8691 & 0.8676\\  \hline
        Bayesian--CNN &0.8848 & 0.8817  & 0.8801\\ \hline
        modified Bayesian-CNN & 0.8898 &0.8856 &0.8837\\ 
        \hline
    \end{tabular}
    \label{Tab:comp1}
\end{table}

\subsection{Quality assessment metrics}
The performance of the classifier is presented using the confusion matrix in Fig.~\ref{fig:conf_comp}. Details of the confusion matrix is provided in the supplementary document. 
\begin{figure}[H]
    \centering
    \hspace{-5pt}
    \subfigure[]{\label{fig:a}\includegraphics[width=0.16\textwidth]{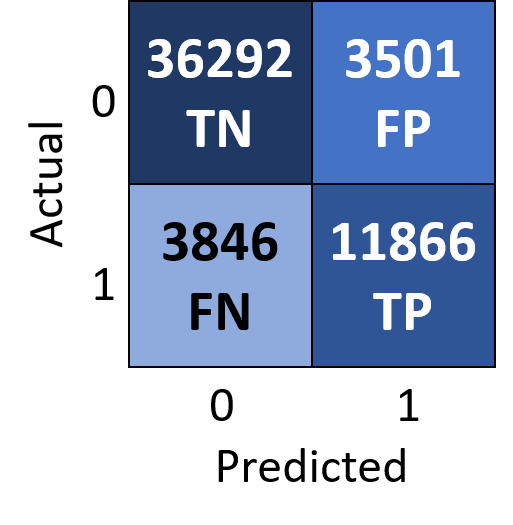}}
    \vspace{-5pt}
    \hspace{-5pt}
    \subfigure[]{\label{fig:b}\includegraphics[width=0.16\textwidth]{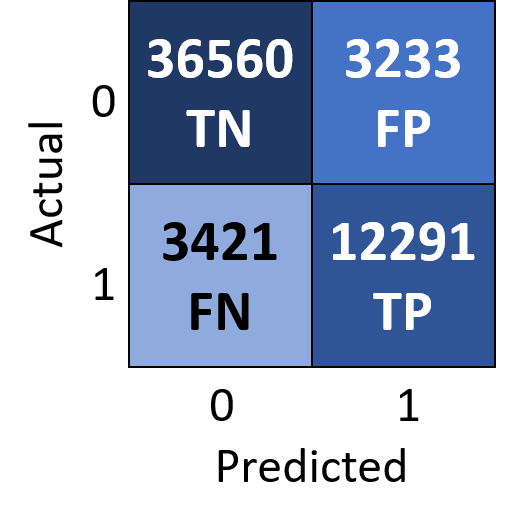}}
    \hspace{-5pt}
    \subfigure[]{\label{fig:b}\includegraphics[width=0.16\textwidth]{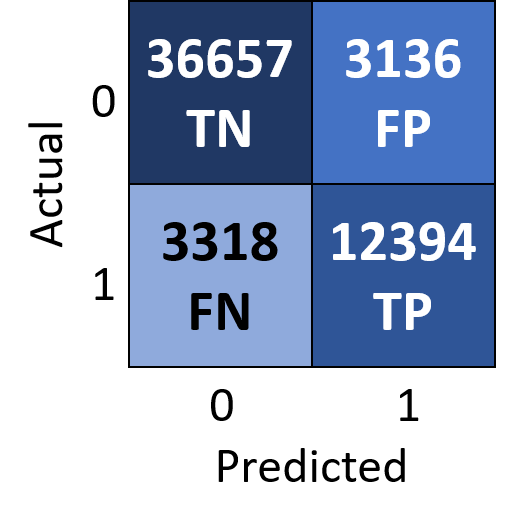}}
    \caption{Comparison of confusion matrices for a) TL-CNN, b) Bayesian--CNN and modified Bayesian--CNN}
    \label{fig:conf_comp}
    \vspace{-12pt}
\end{figure}
In addition to the improvement of accuracy, the modified Bayesian--CNN remarkably reduces the number of false-negative predictions as seen in Fig.~\ref{fig:conf_comp}, which is a significant achievement. CNN is known to perform poorly for biased data set \cite{buda2018systematic}. Given the data set used in this work is highly biased to the negative class, the  TL-CNN model fails to overcome this bias yielding more negative predictions. We demonstrate that the modified Bayesian--CNN, overcomes this limitation of the CNN and provides an unbiased result. The modified Bayesian--CNN reduces the false negative and false positive predictions by about 13.7\% and 10.4\% respectively as compared to TL-CNN. The modified Bayesian--CNN reduces both the false negative and false positive by 3\% as compared to Bayesian--CNN as seen in Fig.~\ref{fig:conf_comp},  demonstrating unbiased prediction. This is a remarkable advancement given that medical imaging data sets are usually biased towards negative labels. This significant improvement in performance is achieved despite having only 1.11 million parameters as compared to 134.33 million for TL-CNN.

 {The improvement in the performance of modified Bayesian--CNN is due to the adaptive nature of the activation function. The adaptive activation function with a probabilistic learning parameter works better because through it the network trains an ensemble of activation functions where each activation function has its weights drawn from a probability distribution which is learned from the data. Learning the probability distribution for the activation parameter from the training data reduces the bias in the model by eliminating the assumption in the functional form of the activation.}

To quantitatively assess the performance of the proposed binary classification the following metrics are used in addition to the confusion matrix: Accuracy, Precision, Recall, F1-score, Cohen's-kappa (CK) coefficient \cite{cohen1960coefficient}. Cohen's Kappa coefficient varies between 0-1 where: 0 – agreement is equivalent to chance and 1.0 – perfect agreement. Improvement in Cohen's Kappa coefficient is achieved by using the modified Bayesian--CNN as seen from table~\ref{Tab:comp2}  which implies that it works better than both Bayesian--CNN and TL-CNN for an unbalanced data set.

The performance of the neural network classifiers is evaluated based on all of the above metrics and the results are provided in table~\ref{Tab:comp2}. The modified Bayesian--CNN performs slightly better than Bayesian--CNN and much better than TL-CNN on all the performance metrics for this data set. 

\begin{table}[H]
    \centering
    \caption{Performance metrics of TL-CNN and BCNN}
    \begin{tabular}{ |p{2.5cm}|p{0.8cm}|p{1.1cm}|p{0.8cm}|p{0.78cm}|} 
        \hline
        \textbf{Network}  & \textbf{Recall} &\textbf{Precision} &\textbf{F1-score} &\textbf{CK}\\
        \hline
        TL-CNN & 0.7552 & 0.7722 &0.7636& 0.6717 \\  \hline
        Bayesian--CNN &0.7823 & 0.7917& 0.7870& 0.7036 \\ \hline
        modified Bayesian--CNN &0.7888 &0.7981& 0.7934 &0.7125 \\ \hline
    \end{tabular}
    \label{Tab:comp2}
\end{table}

\subsubsection{ {Statistical significance}}
 {In order to ensure that the difference in the results obtained by the Bayesian--CNN and the modified Bayesian--CNN is not by chance, we performed a statistical significance test. Importance of statistical significance tests is highlighted in \cite{diong2018poor, rajaraman2020analyzing}. The McNemar's test} \cite{dietterich1998approximate} {is performed and the results show that the difference in performance is statistically significant $(p<0.001)$. Therefore, the Bayesian--CNN and the adaptive Bayesian CNN are statistically different models. The  details and the results of the statistical significance test are provided in Sec. VIII of the supplementary document.}

\subsection{Uncertainty quantification}
\begin{figure}[H]
    \centering
    \includegraphics[width=0.4\textwidth]{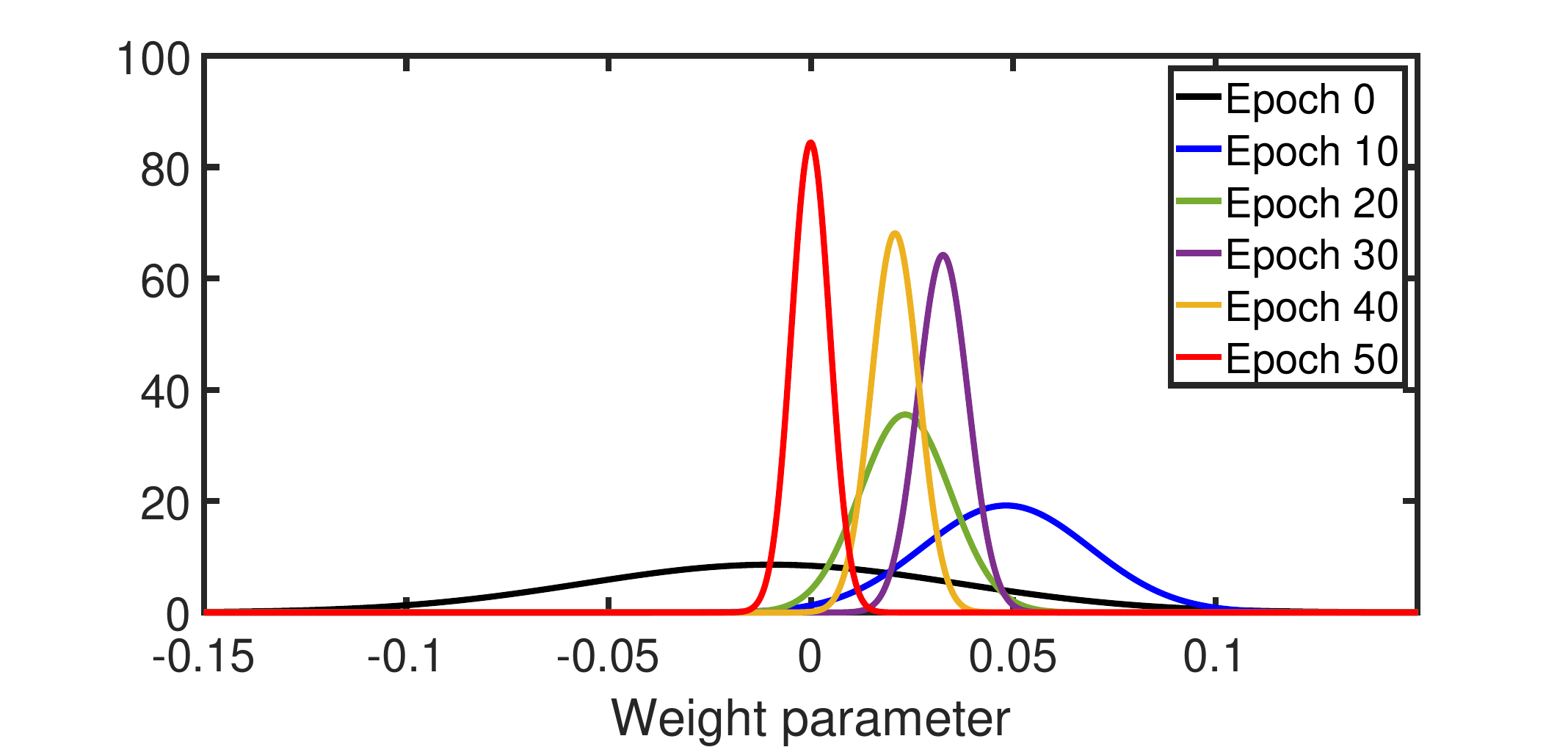}
     \caption{Probability density of a parameter for different epochs of the modified Bayesian-CNN.}
     \label{fig:var_shift}
\end{figure}

Uncertainty quantification informs how confident the network is in its predictions, which is crucial for medical applications.
Thus in addition to the accuracy metrics, uncertainty is an important measure that can be estimated naturally through the Bayesian approaches. On the contrary, the parameters of CNN are deterministic and large in number (134 million) that makes its uncertainty quantification infeasible.
The parameters of the (modified) Bayesian--CNN are probability distributions whose variance provides an estimate of the uncertainties associated with the predictions. The larger the variances the more uncertain the model is on its predictions. The change of the probability distribution of a parameter for the modified Bayesian--CNN is shown for different epochs in Fig.~\ref{fig:var_shift}. 
As the training continues, the standard deviation decreases increasing the confidence in predictions (see the supplementary document).

\begin{figure}[H]
    \centering
    \subfigure[]{\label{fig:neg_a}\includegraphics[scale=1.2]{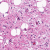}}
    \subfigure[]{\label{fig:neg_b}\includegraphics[scale=1.2]{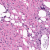}}
    \subfigure[]{\label{fig:neg_c}\includegraphics[scale=1.2]{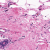}}\\
    \subfigure[]{\label{fig:neg_d}\includegraphics[scale=1.2]{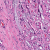}}
    \subfigure[]{\label{fig:neg_e}\includegraphics[scale=1.2]{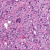}}
    \subfigure[]{\label{fig:neg_f}\includegraphics[scale=1.2]{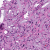}}
    \vspace{-5pt}
    \caption{Sample images having negative labels and their uncertainties:  a) TN [$A=9.25 \times10^{-3}$, $E= 6.50 \times10^{-6}$], b) TN [$A=1.8 \times10^{-2}$, $E=5.65 \times10^{-5}$], c) TN [$A=4.44 \times10^{-3}$, $E=2.53 \times10^{-6}$], d) TN [$A=2.77 \times10^{-1}$, $E=5.42 \times10^{-3}$], e) FP [$A=3.19 \times10^{-1} $, $E=4.59 \times10^{-3}$], f) FP [$A=3.74\times 10^{-1} $, $E=5.57 \times10^{-3} $]. Aleatoric and epistemic uncertainties are given by A and E respectively.}
    \label{fig:exuq1}
    \vspace{-10 pt}
\end{figure}

\begin{figure}[H]
    \centering
    \subfigure[]{\label{fig:pos_a}\includegraphics[scale=1.2]{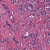}}
    \subfigure[]{\label{fig:pos_b}\includegraphics[scale=1.2]{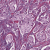}}
    \subfigure[]{\label{fig:pos_c}\includegraphics[scale=1.2]{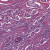}}\\
    \subfigure[]{\label{fig:pos_d}\includegraphics[scale=1.2]{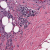}}
    \subfigure[]{\label{fig:pos_e}\includegraphics[scale=1.2]{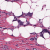}}
    \subfigure[]{\label{fig:pos_f}\includegraphics[scale=1.2]{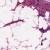}}
    \vspace{-5pt}
    \caption{Sample images having positive labels and their uncertainties: 
    a) TP [$A=7.0 \times10^{-2}$, $E=2.60 \times10^{-4}$], b) TP [$A=6.60\times 10^{-2}$ , $E=1.74 \times10^{-4}$], c) TP [$A=1.14 \times10^{-1}$,  $E=4.91 \times10^{-4}$], d) TP [$ A=4.72 \times10^{-1}$,  $E=2.03 \times10^{-2}$], e) FN [ $A=4.64 \times10^{-1}$, $E=9.37 \times10^{-3}$] f) FN [$A=4.74 \times10^{-1}$, $E=6.26 \times10^{-3}$].}
    \label{fig:exuq2}
    \vspace{-10 pt}
\end{figure}
The uncertainty values associated with an individual image denote the confidence with which the network predicts the class label for that image. The aleatoric uncertainty corresponds to the noise in the data which cannot be reduced by improving the model. The epistemic uncertainty represents the variability due to the model, which can be improved. A set of sample images from the test data set, their predicted class and their aleatoric and epistemic uncertainty in prediction are shown in Fig.~\ref{fig:exuq1} and \ref{fig:exuq2}. The images with negative and positive ground truth labels are shown in Fig.~\ref{fig:exuq1} and Fig.~\ref{fig:exuq2} respectively. The important features apparent from these images are the texture density and color. It appears that the positive class has darker and denser images. It seems that the images with high epistemic uncertainty, Fig.~\ref{fig:exuq1}(d,e,f) and \ref{fig:exuq2}(d,e,f) has  features from both the classes. Therefore, the model gives high epistemic uncertainty to those images indicating mixed features or mislabeling. Thus, the Bayesian approach provides an avenue to identify which images should be referred back to a human expert.

\subsection{Use of uncertainty to improve performance}
\begin{figure}[H]
    \centering
    \subfigure[ {Fraction of test data having uncertainty below the threshold and their accuracy plotted against the uncertainty--threshold.}]{\label{fig:unlim_a}\includegraphics[width=0.5\textwidth]{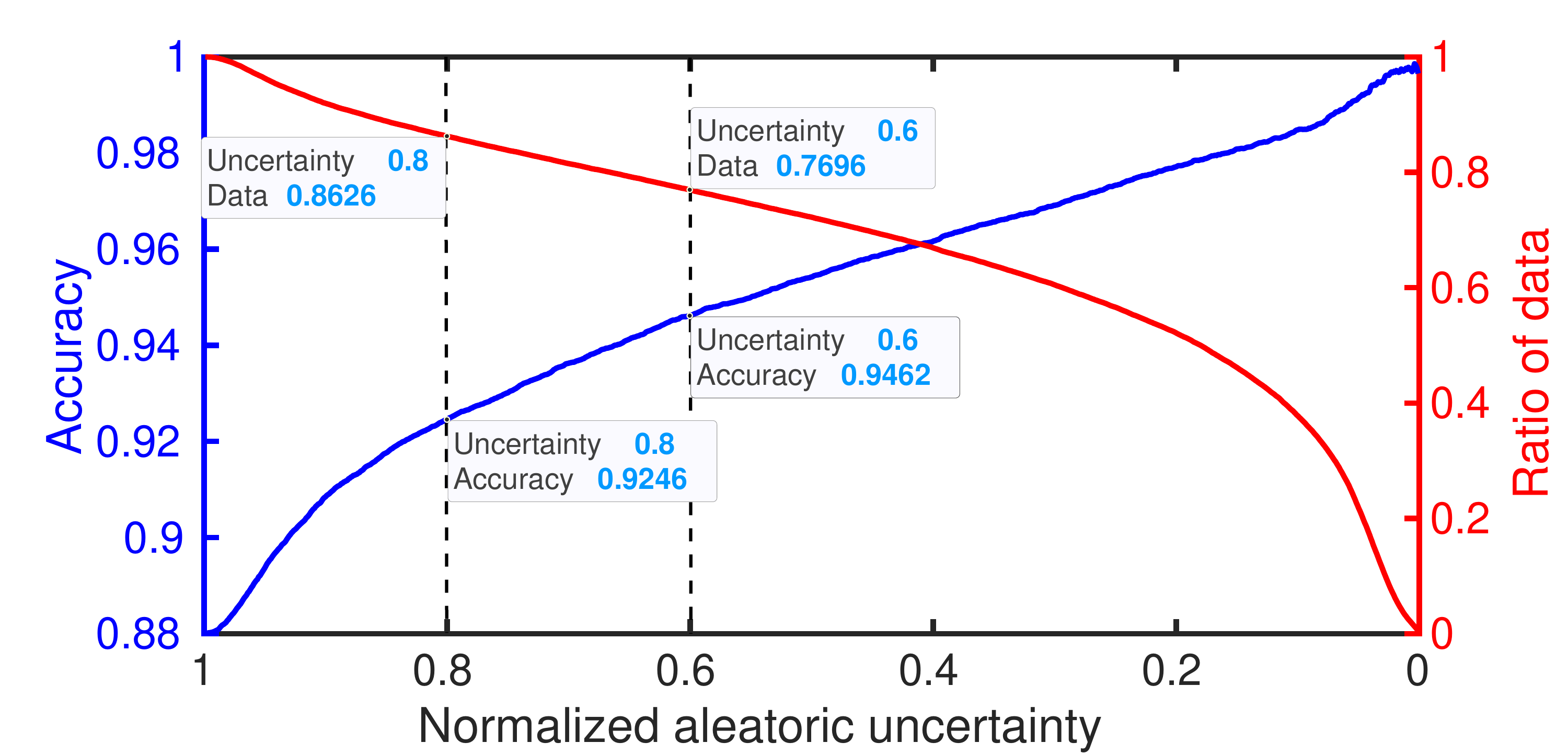}}
    \subfigure[ {Number of false predictions in the low uncertainty subset plotted against the uncertainty--threshold.}]{\label{fig:unlim_b}\includegraphics[width=0.5\textwidth]{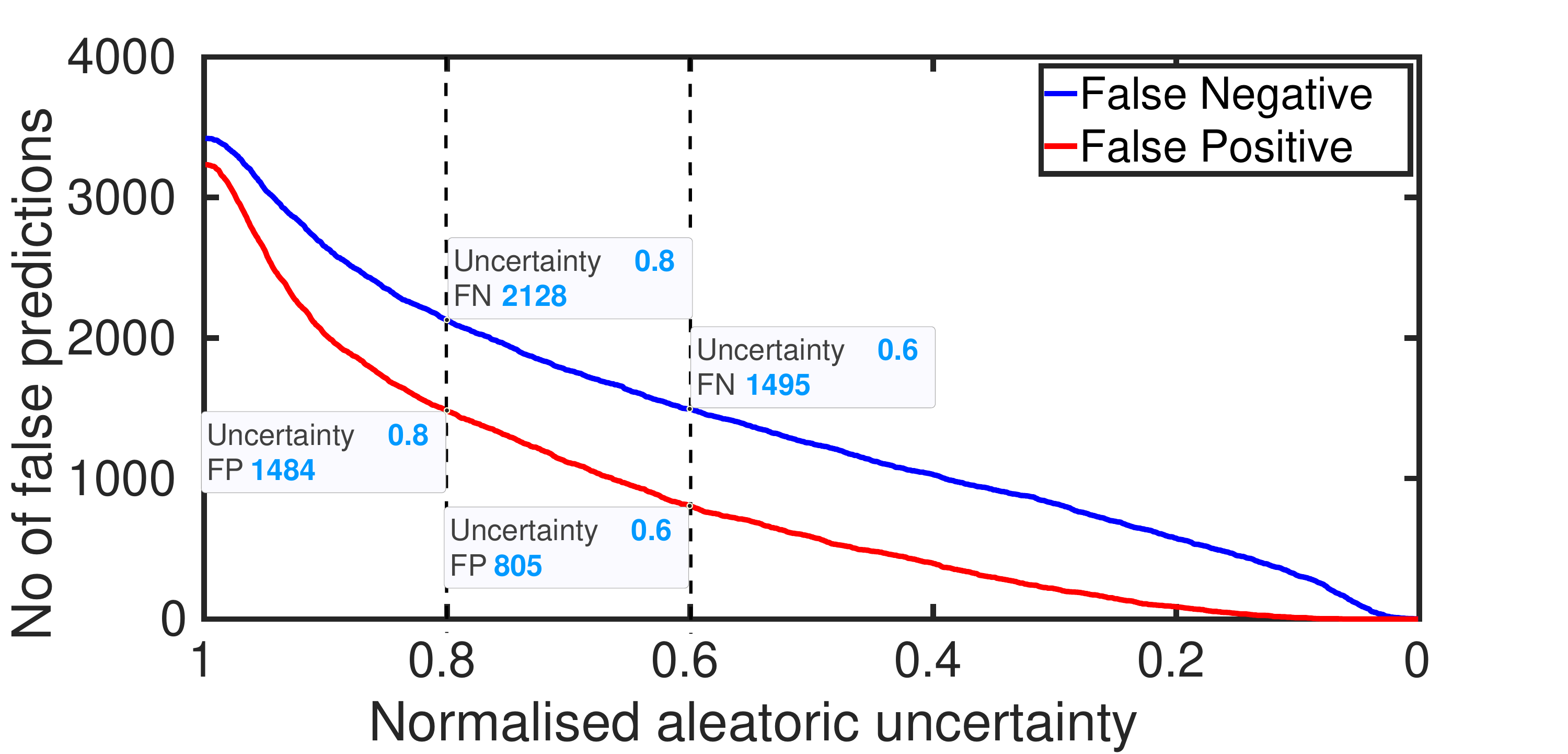}}
    \caption{ {Performance improvements for a subset of test data that has uncertainty less than the uncertainty--threshold. Dotted Lines represents two thresholds of aleatoric uncertainty.}}
    \vspace{-10pt}
    \label{fig:un_limits}
\end{figure}

We utilize the uncertainty to further improve the performance on a subset of data.
 {The aleatoric uncertainty of the test set is normalised by the maximum and minimum values of aleatoric uncertainty of the training set.} For every threshold of aleatoric uncertainty, we divide the test data into two subsets having uncertainty lower and higher than the threshold. The ratio of the number of images in the low uncertainty subset is plotted against the uncertainty--threshold used to create the subset in Fig.~\ref{fig:unlim_a}.  The accuracy corresponding to each of this subset are also plotted against the threshold. The number of false-negative and false-positive predictions are plotted for different thresholds of aleatoric uncertainty in Fig.~\ref{fig:unlim_b}. 
 For instance, if the threshold value is  0.6, the low uncertainty subset contains $77\%$ of the test data and its accuracy is $94.6\%$ which is about $6\%$ improvement in accuracy over the entire test data set. The remaining $23\%$ of the test data, which has uncertainty higher than this threshold (0.6) may be referred to a human expert for a more accurate prediction.  {Furthermore}, for this subset, the percentage of false negatives reduces from $6.2\%$ for the entire data set to $2.7\%$ and the percentage of false positives reduces from $5.8\%$ for the entire data set to $1.5\%$, which are significant improvements.  {We found that normalized uncertainty values lesser than 0.8 provided significant improvement in performance for this data set. For other data sets this threshold on uncertainty can be set by looking at change in slope of the accuracy vs uncertainty plot.}
A similar study is done using the epistemic uncertainty and the results are presented in the supplementary document.
Therefore, we show that uncertainty quantification can be used to improve the performance significantly for a subset of test data.

\subsection{Explanation of uncertainty by visualization in low dimension}
In the following, we provide an explanation of uncertainty by visualizing the test data in a low dimensional space (latent space). A dimensionality reduction is performed on the data using t-SNE to project the 7500-dimensional feature into  3 dimensions (3D). 
\begin{figure}[H]
    \centering
    \vspace{-5pt}
    \subfigure[Low uncertainty]{\label{fig:tsne_low}\includegraphics[width=0.45\textwidth]{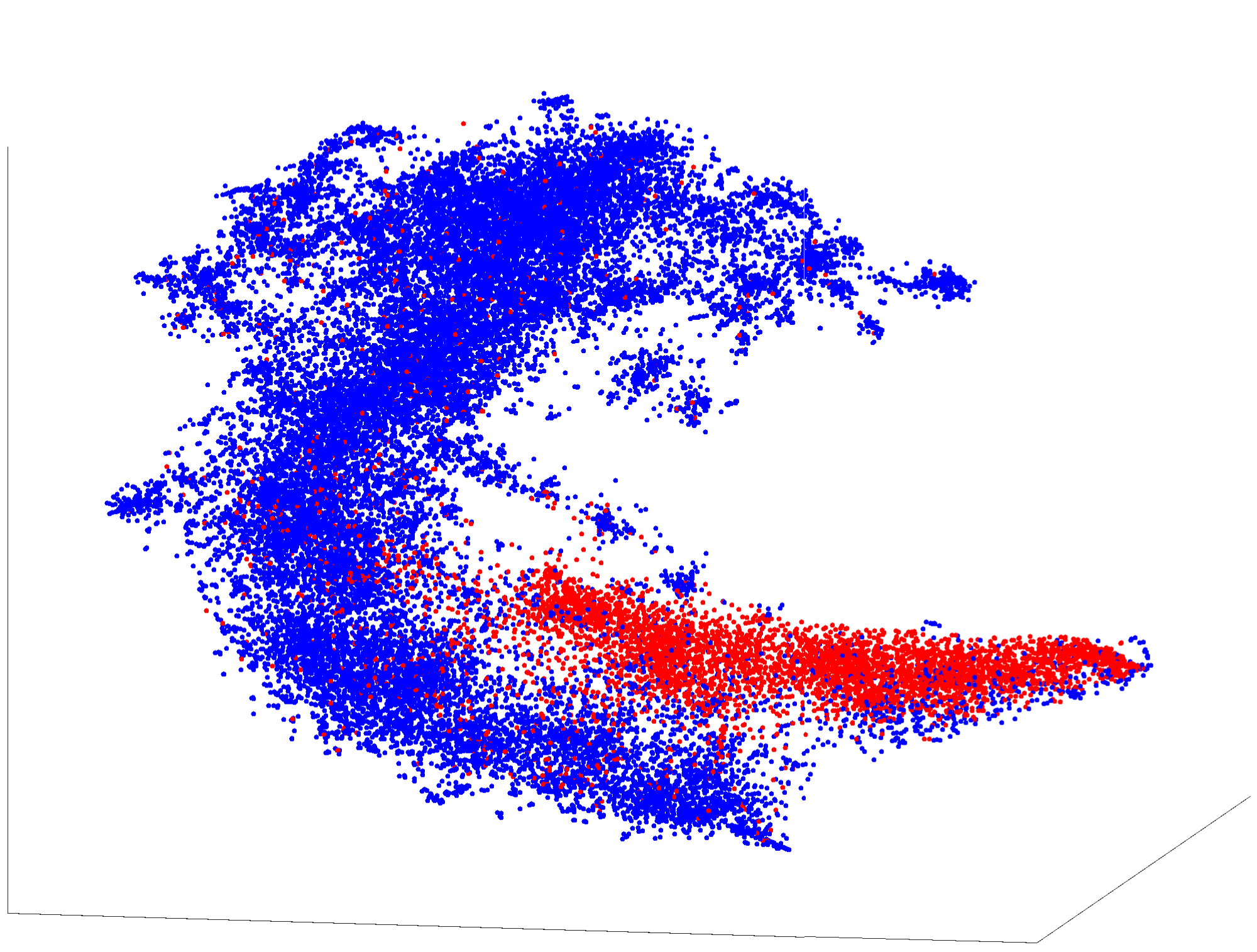}}
    \vspace{-10pt}
    \subfigure[Medium uncertainty]{\label{fig:tsne_med}\includegraphics[width=0.45\textwidth]{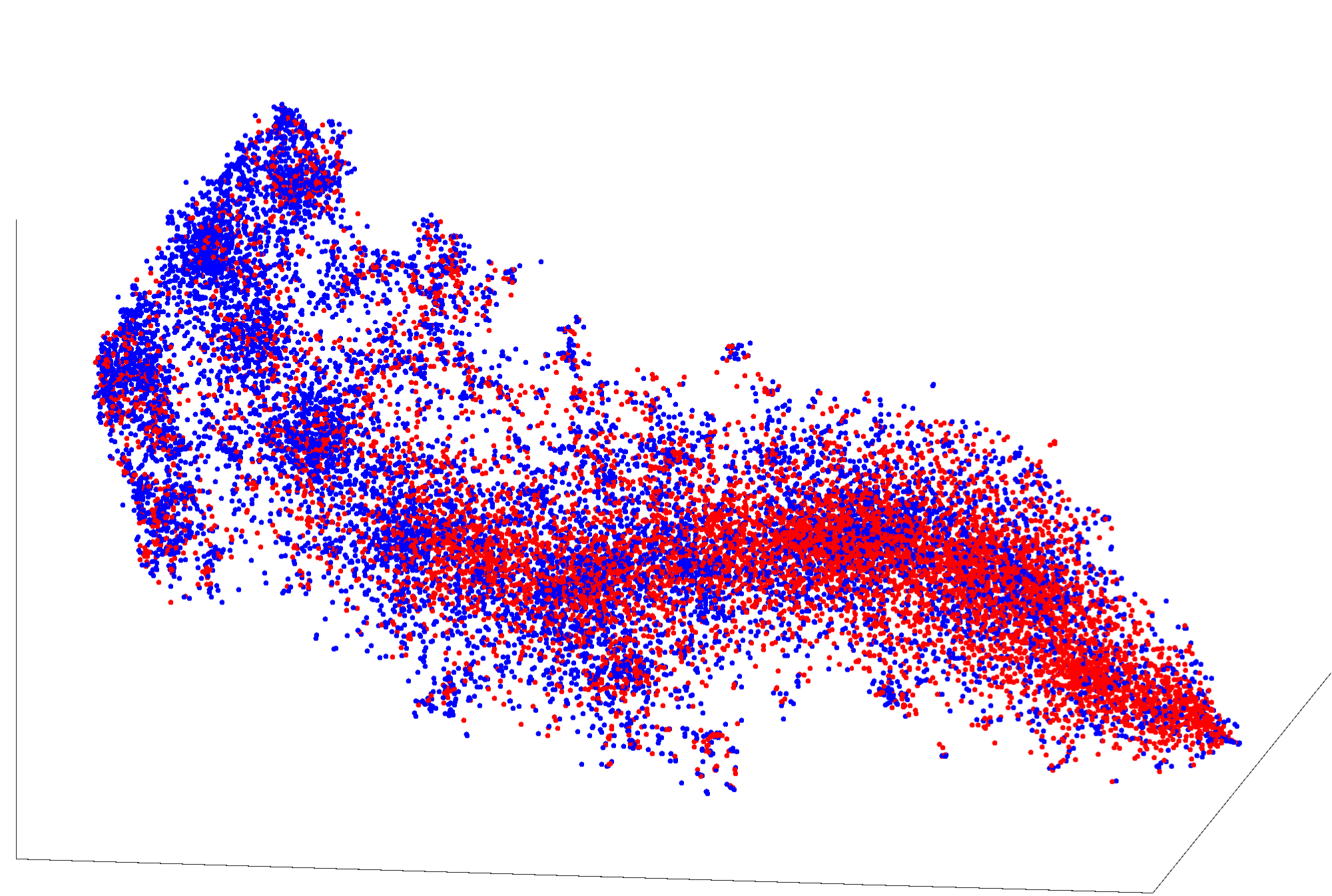}}
    \vspace{-10pt}
    \subfigure[High uncertainty]{\label{fig:tsne_high}\includegraphics[width=0.45\textwidth]{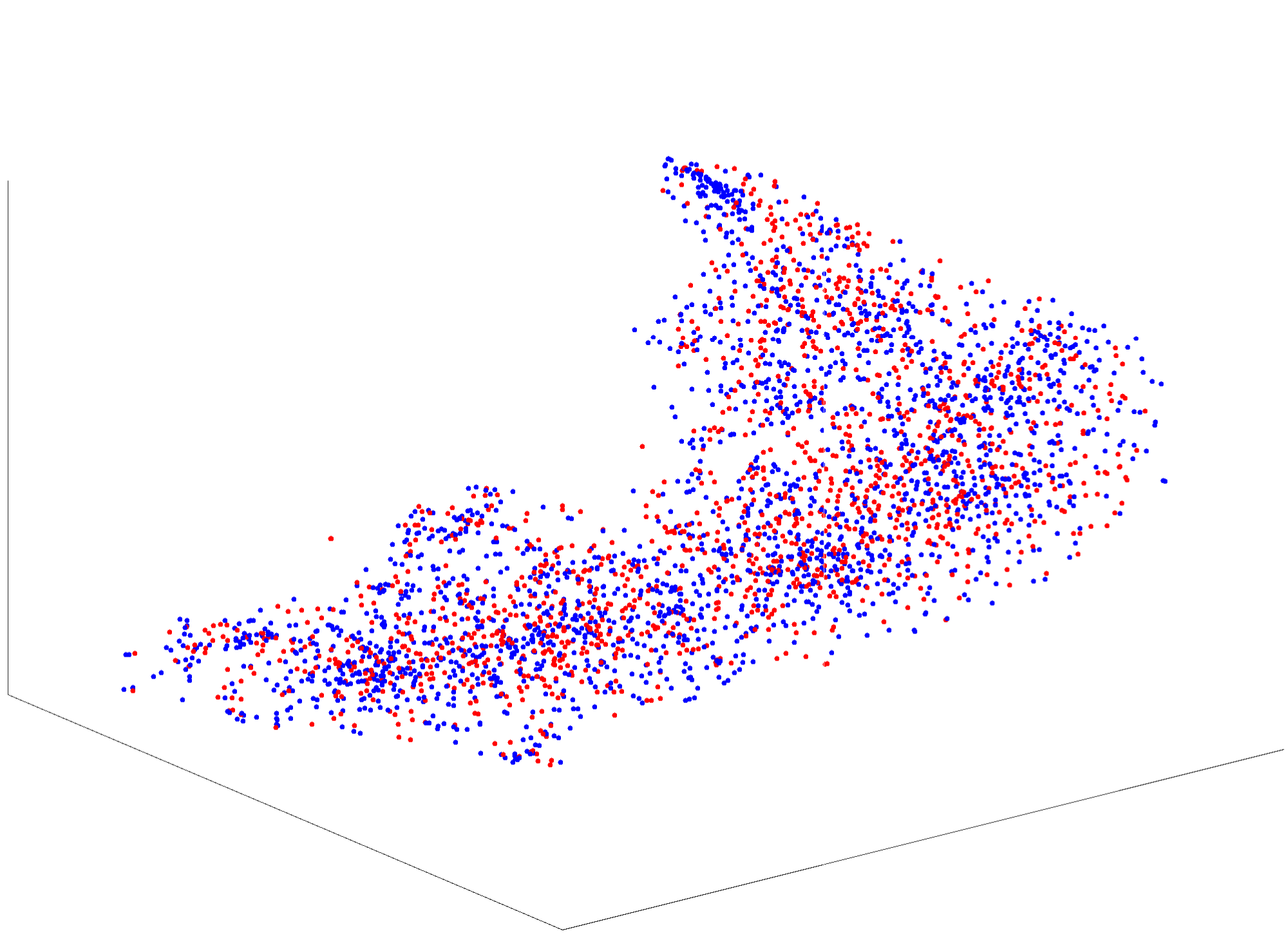}}
\caption{Low dimensional representation of the test data for three subsets using t-SNE. blue: negative class, red: positive class} 
    \label{fig:tsne}
    \vspace{-10 pt}
\end{figure}
The epistemic uncertainty ($E$) evaluated through the Bayesian approach is normalized to range between $[0-1]$. The test data set is divided into three subsets: low ($E \le 0.01$), medium ($0.01<E \le 0.1$) and high uncertainty ($E>0.1$). 
t-SNE is performed for these three subsets. Low dimensional representations obtained via t-SNE are presented in Fig.~\ref{fig:tsne} where each data point represents a test image. Fig.~\ref{fig:tsne} explains the uncertainty associated with the individual images. 
The low uncertainty images have a clear separation between the positive and negative classes in the latent space. Thus the model  {classifies} these images with high confidence. The images with medium uncertainty are clustered at different locations in the latent space however have significant overlap. Therefore the model classifies these images with lesser confidence.
For the high uncertain images, the negative and the positive classes are not at all separable in the latent space. Thus the model cannot distinguish between the two classes for this subset, yielding poor accuracy and low confidence. 


\section{Conclusions}\label{CHAPTER4}
In this study, we have quantified the uncertainty in the classification of histopathological images through Bayesian--CNN and compared its performance against the state-of-the-art.  We have explained the uncertainty and utilized the uncertainty to improve the performance. To conclude this study, we summarize its main findings.

Firstly, we have shown that Bayesian--CNN performs better than the state-of-the-art transfer learning CNN on all performance metrics and provides uncertainties in classifying breast histopathological images. Bayesian--CNN improves the test accuracy by 1.2\% and reduces the false negative and false positive by 11\% and 7.7\% respectively as compare to TL-CNN, despite having only 1.86 million parameters as compared to 134.33 million for TL-CNN.

Secondly, we explained the uncertainties in the test data by projecting them into a low-dimensional space. This low-dimensional projection enables visual interpretation of the data.  The data structure in the low dimensional feature space reveals that images with high uncertainty are not separable whereas the images with low uncertainty are clearly separable. 

Thirdly, using different uncertainty thresholds, we demonstrate that the accuracy of Bayesian networks can be significantly improved on a large fraction of test data (6\% improvement on 77\% test data). 

Fourthly, we have proposed a novel stochastic--adaptive activation enabled Bayesian--CNN and call it modified Bayesian--CNN. In the proposed model we introduce a probabilistic learnable activation function that adapts to the training data to improve predictive capabilities. We have shown that the modified Bayesian-CNN performs slightly better than the Bayesian--CNN in all performance metrics. Especially it can reduce the false negative and false positive predictions by 3\% as compared to Bayesian--CNN. 

Fifthly, we have demonstrated that using the proposed modified Bayesian--CNN and the Bayesian--CNN, the problem of overfitting can be nearly eliminated without any recourse to regularization. That is the Bayesian approach work with almost equal accuracy for unknown and known data sets. 

This work improves the understanding of the uncertainties in Bayesian--CNN-based classification and helps to leverage these uncertainties to further improve its performance.  These findings require further investigation on larger data sets, however, they show potential routes to improve upon the state--of--the--art classifier for a broad range of biomedical applications. 

\bibliographystyle{ieeetr}
\bibliography{References}

\end{document}